\documentclass[11pt,a4paper]{article}
\usepackage[hyperref]{acl2019}
\usepackage{times}
\usepackage{latexsym}
\usepackage{graphicx}
\usepackage{float}

\usepackage{amssymb}

\usepackage{amsmath}
\DeclareMathOperator*{\argmax}{arg\,max}

\usepackage{url}

\aclfinalcopy 
\def\aclpaperid{1997} 

\newcommand*\samethanks[1][\value{footnote}]{\footnotemark[#1]}
\title{Careful Selection of Knowledge to solve Open Book Question Answering}

\author{Pratyay Banerjee\thanks{\quad These authors contributed equally to this work. \href{http://github.com/ari9dam/OBQA}{Github}. } \and  Kuntal Kumar Pal\samethanks \and Arindam Mitra\samethanks \and Chitta Baral 
\\ Department of Computer Science, Arizona State University
\\ \texttt{pbanerj6,kkpal,amitra7,chitta}@asu.edu
}

\date{}
\begin{document}
\maketitle
\begin{abstract}

Open book question answering is a type of natural language based QA (NLQA) where questions are expected to be answered with respect to a given set of open book facts, and common knowledge about a topic. Recently a challenge involving such QA, OpenBookQA, 
has been proposed. Unlike most other NLQA tasks that focus on linguistic understanding, OpenBookQA requires deeper reasoning involving linguistic understanding as well as
reasoning with common knowledge. In this paper we address QA with respect to the OpenBookQA dataset and combine state of the art language models with abductive information retrieval (IR), information gain based re-ranking, passage selection and weighted scoring to achieve 72.0\% accuracy, an 11.6\% improvement over the current state of the art.
\end{abstract}

\section{Introduction}
Natural language based question answering (NLQA) not only involves linguistic understanding, but often involves reasoning with various kinds of knowledge. In recent years, many NLQA datasets and challenges have been proposed, for example, SQuAD \cite{rajpurkar2016squad}, TriviaQA \cite{joshi2017triviaqa} and MultiRC \cite{khashabi2018looking}, and each of them have their own focus, sometimes by design and other times by virtue of their development methodology. Many of these datasets and challenges try to mimic human question answering settings. One such setting is open book question answering where humans are asked to answer questions in a setup where they can refer to books and other materials related to their questions. In such a setting, the focus is not on memorization but, as mentioned in \citet{OpenBookQA2018}, on ``{\em deeper understanding of the materials and its application to new situations \cite{jenkins1995open,jlands1996}.}'' In \citet{OpenBookQA2018}, they propose the OpenBookQA dataset mimicking this setting. 

\begin{table}[H]
\begin{center}
\begin{tabular}{|p{7cm}|}
\hline \textbf{Question:} \textit{A tool used to identify the percent chance of a trait being passed down has how many squares ? } (A) Two squares (B) \textbf{Four squares} (C) Six squares (D) Eight squares \\  
\hline
\textbf{Extracted from OpenBook}:\\
a punnett square is used to identify the percent 
chance of a trait being passed down from a parent 
to its offspring.\\
\hline
\textbf{Retrieved Missing Knowledge}: \\
Two squares is four. \\
The Punnett square is made up of 4 squares and 2 of them are blue and 2 of them are brown, this means you have a 50\% chance of having blue or brown eyes. \\
\hline
\end{tabular}
\end{center}
\caption{\label{tab1} An example of distracting retrieved knowledge }
\end{table}

The OpenBookQA dataset has a collection of questions and four answer choices for each question. The dataset comes with 1326 facts representing an open book. It is expected that answering each question requires at least one of these facts. In addition it requires common knowledge. To obtain relevant common knowledge we use an IR system \cite{clark2016combining} front end to a set of knowledge rich sentences.  Compared to reading comprehension based QA (RCQA) setup where the answers to a question is usually found in the given small paragraph, in the OpenBookQA  setup the open book part is much larger (than a small paragraph) and is not complete as additional common knowledge may be required. This leads to multiple challenges. First, finding the relevant facts in an open book (which is much bigger than the small paragraphs in the RCQA setting) is a challenge. Then, finding the relevant common knowledge using the IR front end is an even bigger challenge, especially since standard IR approaches can be misled by distractions. For example, Table \ref{tab1} shows a sample question from the OpenBookQA dataset. We can see the retrieved missing knowledge contains words which overlap with both answer options A and B. Introduction of such knowledge sentences increases confusion for the question answering model. Finally, reasoning involving both facts from open book, and common knowledge leads to multi-hop reasoning with respect to natural language text, which is also a challenge.



We address the first two challenges and make the following contributions in this paper:
 (a) We improve on knowledge extraction from the OpenBook present in the dataset. We use semantic textual similarity models that are trained with different datasets for this task; (b) We propose natural language abduction to generate queries for retrieving missing knowledge; (c) We show how to use Information Gain based Re-ranking to reduce distractions and remove redundant information; (d) We provide an analysis of the dataset and the limitations of BERT Large model for such a question answering task.
 
 The current best model on the leaderboard of OpenBookQA is the BERT Large model \cite{devlin2018bert}. It has an accuracy of 60.4\% and does not use external knowledge. Our knowledge selection and retrieval techniques achieves an accuracy of 72\%, with a margin of 11.6\% on the current state of the art. We study how the accuracy of the BERT Large model varies with varying number of knowledge facts extracted from the OpenBook and through IR.

\section{Related Work}
In recent years, several datasets have been proposed for natural language question answering \cite{rajpurkar2016squad,joshi2017triviaqa,khashabi2018looking,richardson2013mctest,lai2017race,reddy2018coqa,choi2018quac,tafjord2018quarel,mitra2019declarative} and many attempts have been made to solve these challenges \cite{devlin2018bert,vaswani2017attention,seo2016bidirectional}.

Among these, the closest to our work is the work in \cite{devlin2018bert} which perform QA using fine tuned language model  and the works of \cite{Sun2018ImprovingMR,zhang2018kg} which performs QA using external knowledge. 

Related to our work for extracting missing knowledge are the works of \cite{ni2018learning,musa2018answering,khashabi2017learning} which respectively generate a query either by extracting key terms from a question and an answer option or  by classifying key terms or by Seq2Seq models to generate key terms. In comparison, we generate queries using the question, an answer option and an extracted fact using natural language abduction.
    
The task of natural language abduction for natural language understanding has been studied for a long time \cite{norvig1983frame,norvig1987inference,hobbs2004abduction,hobbs1993interpretation,wilensky1983planning,wilensky2000berkeley,charniak1988logic,charniak1989semantics}. However, such works transform the natural language text to a logical form and then use formal reasoning to perform the abduction. On the contrary, our system performs abduction over natural language text without translating the texts to a logical form.

\section{Approach}
Our approach involves six main modules: \textit{Hypothesis Generation}, \textit{OpenBook Knowledge Extraction}, \textit{Abductive Information Retrieval}, \textit{Information Gain based Re-ranking}, \textit{Passage Selection} and \textit{Question Answering}. A key aspect of our approach is to accurately hunt the needed knowledge facts from the OpenBook knowledge corpus and hunt missing common knowledge using IR. We explain our approach in the example given in Table \ref{tabgen}.

\begin{table}[H]
\begin{center}
\begin{tabular}{|p{7cm}|}
\hline \textbf{Question:} \textit{A red-tailed hawk is searching for prey. It is most likely to swoop down on what?} (A)  \textbf{a gecko} \\  
\hline
\textbf{Generated Hypothesis} :\\
  H : A red-tailed hawk is searching for prey. It is most likely to swoop down on a gecko. \\
\hline
\textbf{Retrieved Fact from OpenBook:} \\
  F : hawks eat lizards \\
\hline
\textbf{Abduced  Query to find missing knowledge:} \\
  K : gecko is lizard \\
  \hline
  \textbf{Retrieved Missing Knowledge using IR:} \\
  K : Every gecko is a lizard.\\
  \hline
\end{tabular}
\end{center}
\caption{\label{tabgen} Our approach with an example for the correct option}
\end{table}

 
In \textit{Hypothesis Generation}, our system generates a hypothesis $\mathbf{H_{ij}}$ for the $i$th question and $j$th answer option, where $j \in \{1,2,3,4\} $. In \textit{OpenBook Knowledge Extraction}, our system retrieves appropriate knowledge $\mathbf{F_{ij}}$ for a given hypothesis $\mathbf{H_{ij}}$ using semantic textual similarity, from the OpenBook knowledge corpus $\mathbf{F}$. In \textit{Abductive Information Retrieval}, our system abduces missing knowledge from $\mathbf{H_{ij}}$ and $\mathbf{F_{ij}}$. The system formulates queries to perform IR to retrieve missing knowledge $\mathbf{K_{ij}}$.  With the retrieved $\mathbf{K_{ij}}$, $\mathbf{F_{ij}}$, \textit{Information Gain based Re-ranking} and \textit{Passage Selection} our system creates a knowledge passage $\mathbf{P_{ij}}$. In \textit{Question Answering}, our system uses $\mathbf{P_{ij}}$ to answer the questions using a BERT Large based MCQ model, similar to its use in solving SWAG \cite{zellers2018swag}.

\begin{figure}
  \includegraphics[width=7.5cm]{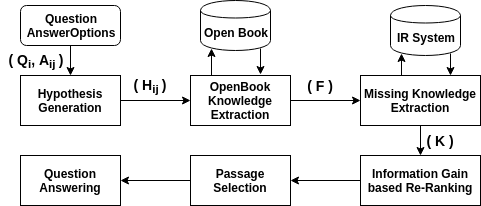}
  \caption{Our approach}
  \label{fig:blk}
\end{figure}



\subsection{Hypothesis Generation}
Our system creates a hypothesis for each of the questions and candidate answer options as part of the data preparation phase as shown in the example in Table \ref{tabgen}. The questions in the OpenBookQA
dataset are either with \textit{wh} word or are incomplete statements.
To create hypothesis statements for questions with \textit{wh} words, we use the rule-based model of \citet{demszky2018transforming}. For the rest of the questions, we concatenate the questions with each of the answers to produce the four hypotheses. This has been done for all the training, test and validation sets. 
\subsection{OpenBook Knowledge Extraction}
To retrieve a small set of relevant knowledge facts from the knowledge corpus $\mathbf{F}$, a textual similarity model is trained in a supervised fashion on two different datasets and the results are compared. We use the \textit{large-cased} BERT \cite{devlin2018bert} (BERT Large) as the textual similarity model. 
\subsubsection{BERT Model Trained on STS-B}
\label{sts}
We train it on the semantic textual similarity (STS-B) data from the GLUE dataset \cite{wang2018glue}. The trained model is then used to retrieve the top ten knowledge facts from corpus $\mathbf{F}$ based on the STS-B scores. The STS-B scores range from 0 to 5.0, with 0 being least similar.

\subsubsection{BERT Model Trained on OpenBookQA}
\label{bbc}
 We generate the dataset using the gold OpenBookQA facts from $\mathbf{F}$ for the train and validation set provided. To prepare the train set, we first find the similarity of the OpenBook $\mathbf{F}$ facts with respect to each other using the BERT model trained on STS-B dataset. We assign a score 5.0 for the gold $\mathbf{\hat{F_i}}$ fact for a hypothesis. We then sample different facts from the OpenBook and assign the STS-B similarity scores between the sampled fact and the gold fact $\mathbf{\mathbf{\hat{F}_{i}}}$ as the target score for that fact $\mathbf{F_{ij}}$ and $\mathbf{H_{ij}}$. For example:\\

 \noindent
 \fbox{\begin{minipage}{19em} 
 \textbf{Hypothesis} : Frilled sharks and angler fish live far beneath the surface of the ocean, which is why they are known as Deep sea animals. \\
 \textbf{Gold Fact} : deep sea animals live deep in the ocean : Score : 5.0 \\
 \textbf{Sampled Facts} : \\
 coral lives in the ocean  : Score : 3.4 \\
 a fish lives in water : Score : 2.8
 \end{minipage}}
 \\
 
 We do this to ensure a balanced target score is present for each hypothesis and fact. We use this trained model to retrieve top ten relevant facts for each $\mathbf{H_{ij}}$ from the knowledge corpus $\mathbf{F}$. 

\subsection{Natural Language Abduction and IR}
To search for the missing knowledge, we need to know what we are missing. We use ``\textit{abduction}'' to figure that out.
Abduction is a long studied task in AI, where normally, both the observation (hypothesis) and the domain knowledge (known fact) is represented in a formal language from which a logical solver abduces possible explanations (missing knowledge). However, in our case, both the observation and the domain knowledge are given as natural language sentences from which we want to find out a possible missing knowledge, which we will then hunt using IR. For example, one of the hypothesis $\mathbf{H_{ij}}$ is ``\textit{A red-tailed hawk is searching for prey. It is most likely to swoop down on a gecko.}'', and for which the known fact $\mathbf{F_{ij}}$ is ``\textit{hawks eats lizards}''. From this we expect the output of the natural language abduction system to be $\mathbf{K_{ij}}$ or ``\textit{gecko is a lizard}''. We will refer to this as ``\textit{natural language abduction}''.


For natural language abduction, we propose three models, compare them against a baseline model and evaluate each on a downstream question answering task. All the models ignore stop words except the Seq2Seq model. We describe the three models and a baseline model in the subsequent subsections.

\subsubsection{Word Symmetric Difference Model}
We design a simple heuristic based model defined as below:
\begin{equation*}
    K_{ij} = (H_{ij} \cup F_{ij}) \setminus (H_{ij} \cap F_{ij}) \quad \forall j \in \{1,2,3,4\}  
\end{equation*}
where $i$ is the $i$th question, $j$ is the $j$th option, $H_{ij}$, $F_{ij}$, $K_{ij}$ represents set of unique words of each instance of hypothesis, facts retrieved from knowledge corpus $\mathbf{F}$ and abduced missing knowledge of validation and test data respectively.
\subsubsection{Supervised Bag of Words Model}
In the Supervised Bag of Words model, we select words which satisfy the following condition: 
\begin{equation*}
    P(w_n \in K_{ij}) > \theta 
\end{equation*}
where $w_n \in \{H_{ij} \cup F_{ij}\}$. To elaborate, we learn the probability of a given word $w_n$ from the set of words in $H_{ij} \cup F_{ij}$ belonging to the abduced missing knowledge $K_{ij}$. We select those words which are above the threshold $\theta$.

To learn this probability, we create a training and validation dataset where the words similar (cosine similarity using spaCy) \cite{spacy2} to the words in the gold missing knowledge $\hat{K}_i$ (provided in the dataset) are labelled as positive class and all the other words not present in $\hat{K}_i$ but in $H_{ij} \cup F_{ij}$ are labelled as negative class.  Both classes are ensured to be balanced. Finally, we train a binary classifier using BERT Large with one additional feed forward network for classification. We define value for the threshold $\theta$ using the accuracy of the classifier on validation set. $0.4$ was selected as the threshold.

\subsubsection{Copynet Seq2Seq Model}
In the final approach, we used the copynet sequence to sequence model \cite{P16-1154} to generate, instead of predict, the missing knowledge given, the hypothesis $\mathbf{H}$ and knowledge fact from the corpus $\mathbf{F}$. The intuition behind using copynet model is to make use of the copy mechanism to generate essential yet \textit{precise} (minimizing distractors) information which can help in answering the question.
We generate the training and validation dataset using the gold $\mathbf{\hat{K}_i}$ as the target sentence, but we replace out-of-vocabulary words from the target with words similar (cosine similarity using spaCy) \cite{spacy2} to the words present in $H_{ij} \cup F_{ij}$. Here, however, we did not remove the stopwords.
We choose one, out of multiple generated knowledge based on our model which provided maximum \textit{overlap\_score}, given by
\begin{equation*}
overlap\_score = \frac{\sum_{i}{count ((\hat{H}_{i} \cup F_{i})\cap K_{i})}}{\sum_{i}{count(\hat{K_{i}})}}
\end{equation*}
where $i$ is the $i$th question, $\hat{H}_{i}$ being the set of unique words of correct hypothesis, $F_{i}$ being the set of unique words from retrieved facts from knowledge corpus $\mathbf{F}$, $K_{i}$ being the set of unique words of predicted missing knowledge and $\hat{K_i}$ being the set of unique words of the gold missing knowledge .


\subsubsection{Word Union Model}

To see if abduction helps, we compare the above models with a Word Union Model.
To extract the candidate words for missing knowledge, we used the set of unique words from both the hypothesis and OpenBook knowledge as candidate keywords. The model can be formally represented with the following:
\begin{equation*}
    K_{ij} = (H_{ij} \cup F_{ij}) \quad \forall j \in \{1,2,3,4\}
\end{equation*}

\subsection{Information Gain based Re-ranking}
In our experiments we observe that, BERT QA model gives a higher score if similar sentences are repeated, leading to wrong classification. Thus, we introduce Information Gain based Re-ranking to remove redundant information.

We use the same BERT Knowledge Extraction model Trained on OpenBookQA data (section \ref{bbc}), which is used for extraction of knowledge facts from corpus $\mathbf{F}$ to do an initial ranking of the retrieved missing knowledge $\mathbf{K}$. The scores of this knowledge extraction model is used as relevancy score, $rel$. To extract the top ten missing knowledge $\mathbf{K}$, we define a redundancy score, $red_{ij}$ , as the maximum cosine similarity, $sim$, between the previously selected missing knowledge, in the previous iterations till $i$, and the candidate missing knowledge $K_j$. 
If the last selected missing knowledge is $K_i$, then
\begin{equation*}
    red_{ij}(K_j) = max(red_{i-1,j}(K_j), sim(K_i,K_j))
\end{equation*}
\begin{equation*}
    rank\_score  = (1-red_{i,j}(K_j))*rel(K_j)
\end{equation*}

For missing knowledge selection, we first take the missing knowledge with the highest $rel$ score. From the subsequent iteration, we compute the redundancy score with the last selected missing knowledge for each of the candidates and then rank them using the updated $rank\_score$. We select the top ten missing knowledge for each $\mathbf{H_{ij}}$. 

\subsection{Question Answering}
Once the OpenBook knowledge facts $\mathbf{F}$ and missing knowledge $\mathbf{K}$ have been extracted, we move onto the task of answering the questions. 

\subsubsection{Question-Answering Model}
We use BERT Large model for the question answering task. For each question, we create a passage using the extracted facts and missing knowledge and fine-tune the BERT Large model for the QA task with one additional feed-forward layer for classification. The passages for the train dataset were prepared using the knowledge corpus facts, $\mathbf{F}$. We create a passage using  the top N  facts, similar to the actual gold fact $\mathbf{\hat{F}_i}$, for the train set. The similarities were scored using the STS-B trained model (section \ref{sts}). The passages for the training dataset do not use the gold missing knowledge $\mathbf{\hat{K}_i}$ provided in the dataset. For each of our experiments, we use the same trained model, with passages from different IR models.

The BERT Large model limits passage length to be lesser than equal to 512. This restricts the size of the passage. To be within the restrictions we create a passage for each of the answer options, and score for all answer options against each passage. We refer to this scoring as \textit{sum score}, defined as follows:

For each answer options, $A_j$, we create a passage $P_j$ and score against each of the  answer options $A_i$. To compute the final score for the answer, we sum up each individual scores. If $Q$ is the question, the score for the answer is defined as
\begin{equation*}
    Pr(Q,A_i) = \sum_{j=1}^{4}score(P_j,Q,A_i)
\end{equation*}
where $score$ is the classification score given by the BERT Large model.
The final answer is chosen based on, 
\begin{equation*}
A = \argmax_A Pr(Q,A_i)
\end{equation*}

\begin{table*}[!t]
\centering
\begin{tabular}{|c|ccc|ccc|ccc|}
\hline
$\mathbf{F}$& \multicolumn{3}{c|}{\textbf{Any Passage}} & \multicolumn{3}{c|}{ \textbf{Correct Passage}} & \multicolumn{3}{c|}{\textbf{Accuracy(\%)}} \\
\hline
\textbf{N} &\textbf{TF-IDF}&\textbf{Trained}&\textbf{STS-B}&\textbf{TF-IDF}&\textbf{Trained}&\textbf{STS-B}&\textbf{TF-IDF}&\textbf{Trained}&\textbf{STS-B}\\
\hline
1&228&258&288&196&229&234&52.6&63.6&59.2\\
2&294&324&347&264&293&304&57.4&\textbf{66.2}&60.6\\
3&324&358&368&290&328&337&59.2&65.0&60.2\\
5&350&391&398&319&\textbf{370}&366&61.6&65.4&62.8\\
7&356&411&411&328&\textbf{390}&384&59.4&65.2&61.8\\
10&373&423&420&354&\textbf{405}&396&60.4&65.2&59.4\\
\hline
\end{tabular}
\caption{Compares (a) The number of correct facts that appears across any four passages (b) The number of correct facts that appears in the passage of the correct hypothesis (c) The accuracy for TF-IDF, BERT model trained on STS-B dataset and BERT model trained on OpenBook dataset. N is the number of facts considered.
  }
  \label{exp:f1}
\end{table*}
\subsubsection{Passage Selection and Weighted Scoring}
In the first round,  we score each of the answer options using a passage created from the selected knowledge facts from corpus $\mathbf{F}$. For each question, we ignore the passages of the answer options which are in the bottom two. We refer to this as \textit{Passage Selection}. In the second round, we score for only those passages which are selected after adding the missing knowledge $\mathbf{K}$.

We assume that the correct answer has the highest score in each round. Therefore we multiply the scores obtained after both rounds. We refer to this as \textit{Weighted Scoring}.
We define the combined passage selected scores and weighted scores as follows :
\begin{equation*}
    Pr(\mathbf{F},Q,A_i) = \sum_{j=1}^{4}{score(P_j,Q,A_i)}
\end{equation*} where $P_j$ is the passage created from extracted OpenBook knowledge, \textbf{F}. The top two passages were selected based on the scores of  $Pr(\mathbf{F},Q,A_i)$. 
\begin{equation*}
    Pr(\mathbf{F}\cup \mathbf{K},Q,A_i) = \sum_{k=1}^{4}{\delta * score(P_k,Q,A_i)}
\end{equation*}
where $\delta=1$ for the top two scores and $\delta=0$ for the rest. $P_k$ is the passage created using both the facts and missing knowledge. The final weighted score is :
\begin{equation*}
        wPr(Q,A_i) = Pr(\mathbf{F},Q,A_i) * Pr(\mathbf{F} \cup \mathbf{K},Q,A_i)
\end{equation*}
The answer is chosen based on the top weighted scores as below:
\begin{equation*}
A = \argmax_A  wPr(Q,A_i)
\end{equation*}

\section{Experiments}

\subsection{Dataset and Experimental Setup}
The dataset of OpenBookQA contains 4957 questions in the train set and 500 multiple choice questions in validation and test respectively.
We train a BERT Large based QA model using the top ten knowledge facts from the corpus $\mathbf{F}$, as a passage for both training and validation set. We select the model which gives the best score for the validation set. The same model is used to score the validation and test set with different passages derived from different methods of Abductive IR. The best Abductive IR model, the number of facts from $\mathbf{F}$ and $\mathbf{K}$ are selected from the best validation scores for the QA task.  

\subsection{OpenBook Knowledge Extraction}
\noindent
\fbox{\begin{minipage}{19em} 
\textbf{Question:}
\textit{.. they decide the best way to save money is ?}
 (A) to quit eating lunch out (B) to make more phone calls (C) to buy less with monopoly money (D) to have lunch with friends\\
 \textbf{Knowledge extraction trained with STS-B:}\\
using less resources usually causes money to be saved \\
a disperser disperses\\
each season occurs once per year\\
 \textbf{Knowledge extraction trained with OpenBookQA:}\\
 using less resources usually causes money to be saved\\
 decreasing something negative has a positive impact on a thing\\
 conserving resources has a positive impact on the environment
\end{minipage}}\\

Table \ref{exp:f1} shows a comparative study of our three approaches for OpenBook knowledge extraction. We show, the number of correct OpenBook knowledge extracted for all of the four answer options  using the three approaches TF-IDF, BERT model trained on STS-B data and BERT model Trained on OpenBook data. Apart from that, we also show the count of the number of facts present precisely across the correct answer options. It can be seen that the Precision@N for the BERT model trained on OpenBook data is better than the other models as N increases. 

The above example presents the facts retrieved from BERT model trained on OpenBook which  are more relevant than the facts retrieved from BERT model trained on STS-B.  Both the models were able to find the most relevant fact, but the other facts for STS-B model introduce more distractors and have lesser relevance. The impact of this is visible  from the accuracy scores for the QA task in Table \ref{exp:f1} . The best performance of the BERT QA model can be seen to be 66.2\% using only OpenBook facts.

\subsection{Abductive Information Retrieval}
\begin{figure}[h!]
  \includegraphics[width=19em]{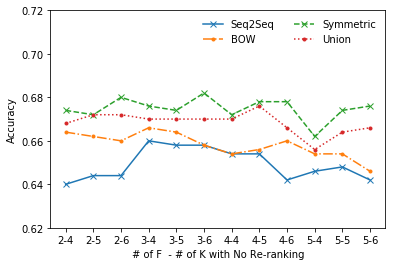}
  \caption{Accuracy v/s Number of facts from $\mathbf{F}$ - number of facts from $\mathbf{K}$, without Information Gain based Re-ranking for 3  abductive IR models and Word Union model. \footnotemark}
  \label{fig:plt1}
\end{figure}

We evaluate the abductive IR techniques at different values for number of facts from $\mathbf{F}$ and number of missing knowledge $\mathbf{K}$ extracted using IR. Figure \ref{fig:plt1} shows the accuracy against different combinations of $\mathbf{F}$ and $\mathbf{K}$ , for all four techniques of IR prior to Information gain based Re-ranking. In general, we noticed that the trained models performed poorly compared to the baselines. The Word Symmetric Difference  model performs better, indicating abductive IR helps. The poor performance of the trained models can be attributed to the challenge of learning abductive inference. 

\begin{figure}[h!]
  \includegraphics[width=19em]{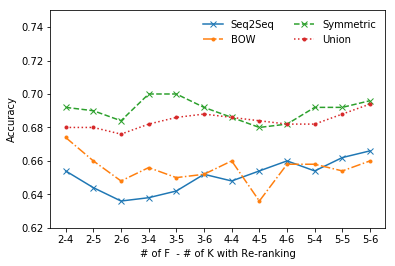}
  \caption{Accuracy v/s Number of facts from $\mathbf{F}$ - number of facts from $\mathbf{K}$, with Information Gain based Re-ranking for 3 abductive IR models and Word Union model. \footnotemark[1]}
  \label{fig:plt2}
\end{figure}

For the above example it can be seen, the pre-reranking facts are relevant to the question but contribute very less considering the knowledge facts retrieved from the corpus $\mathbf{F}$ and the correct answer.
Figure \ref{fig:plt2} shows the impact of Information gain based Re-ranking. Removal of redundant data allows the scope of more relevant information being present in the Top N retrieved missing knowledge $\mathbf{K}$. \\

\noindent
\fbox{\begin{minipage}{19em}
    \textbf{Question}:
 \textit{A red-tailed hawk is searching for prey. It is most likely to swoop down on what?} (A) an eagle (B) a cow (C) \textbf{a gecko} (D) a deer\\
 \textbf{Fact from} $\mathbf{F}$ : hawks eats lizards \\
 \textbf{Pre-Reranking $\mathbf{K}$} :\\
 red-tail hawk in their search for prey\\
 Red-tailed hawks soar over the prairie and woodlands in search of prey.\\
 \textbf{Post-Reranking $\mathbf{K}$}: \\
 Geckos - only vocal lizards. \\
 Every gecko is a lizard. 
\end{minipage}} \\

\subsection{Question Answering}

Table \ref{exp:finalresults} shows the incremental improvement on the baselines after inclusion of carefully selected knowledge.
\begin{table}[t!]
\begin{tabular}{|lr|}
\hline
\textbf{Solver} & \textbf{Accuracy} (\%)\\
\hline
\textit{Leaderboard} & \\
Guess All (``random") & 25.0\\
Plausible Answer Detector & 49.6\\
Odd-one-out Solver & 50.2\\
Question Match & 50.2\\
Reading Strategies & 55.8\\
\hline
\textit{Model - BERT-Large (SOTA)} & \\
Only Question (No KB) & 60.4 \\
\hline
\textit{Model - BERT-Large (Our)} &  \\
$\mathbf{F}$ - TF-IDF& 61.6\\
$\mathbf{F}$ - Trained KE & 66.2\\
$\mathbf{F} \cup \mathbf{K}$& 70.0\\
$\mathbf{F} \cup \mathbf{K}$ with Weighted Scoring& 70.4\\
$\mathbf{F} \cup \mathbf{K}$ with Passage Selection& 70.8\\
$\mathbf{F} \cup \mathbf{K}$ with Both& \textbf{72.0}\\
\hline
\textit{Oracle - BERT-Large} &  \\
$\mathbf{F}$ gold & 74.4\\
$\mathbf{F} \cup \mathbf{K}$ gold & 92.0\\
\hline
\end{tabular}
\caption{Test Set Comparison of Different Components. Current state of the art (SOTA) is the Only Question model. K is retrieved from Symmetric Difference Model. KE refers to Knowledge Extraction.}
  \label{exp:finalresults}
\end{table}


Passage Selection and Weighted Scoring are used to overcome the challenge of boosted prediction scores due to cascading effect of errors in each stage. \\

\noindent
\fbox{\begin{minipage}{19em}
\textbf{Question}:\textit{ What eat plants?} 
 (A) leopards (B) eagles (C) owls (D) \textbf{robin} \\ 
 \textbf{Appropriate extracted Fact from $\mathbf{F}$} :\\
 some birds eat plants \\
 \textbf{Wrong Extracted Fact from $\mathbf{F}$}  :\\
 a salamander eats insects  \\
 \textbf{Wrong Retrieved Missing Knowledge}: \\
 Leopard geckos eat mostly insects 
\end{minipage}} \\

For the example shown above, the wrong answer \textit{leopards} had very low score with only the facts extracted from knowledge corpus $\mathbf{F}$. But introduction of missing knowledge from the wrong fact from $\mathbf{F}$  boosts its scores, leading to wrong prediction. Passage selection helps in removal of such options and Weighted Scoring gives preference to those answer options whose scores are relatively high before and after inclusion of missing knowledge.

\footnotetext{No Passage Selection and Weighted Scoring.}

\section{Analysis \& Discussion}
\subsection{Model Analysis}

\textbf{BERT Question Answering model}:
BERT performs well on this task, but is prone to distractions. Repetition of information leads to boosted prediction scores. BERT performs well for lookup based QA, as in RCQA tasks like SQuAD. But this poses a challenge for Open Domain QA, as the extracted knowledge enables lookup for all answer options, leading to an adversarial setting for lookup based QA. This model is able to find the correct answer, even under the adversarial setting, which is shown by the performance of the \textit{sum score} to select the answer after passage selection.
     
    \textbf{Symmetric Difference Model}
     This model improves on the baseline Word Union model by 1-2\%. The improvement is dwarfed because of inappropriate domain knowledge from $\mathbf{F}$ being used for abduction. The intersection between the inappropriate domain knowledge and the answer hypothesis is $\mathbf{\varnothing}$, which leads to queries which are exactly same as the Word Union model.
    
    \textbf{Supervised learned models}
     The supervised learned models for abduction under-perform. The Bag of Words and the Seq2Seq models fail to extract keywords for many $\mathbf{F}-\mathbf{H}$ pairs, sometimes missing the keywords from the answers. The Seq2Seq model sometimes extracts the exact missing knowledge, for example it generates ``\textit{some birds is robin}'' or ``\textit{lizard is gecko}''. This shows there is promise in this approach and the poor performance can be attributed to insufficient train data size, which was 4957 only. A fact verification model might improve the accuracy of the supervised learned models. But, for many questions, it fails to extract proper keywords, copying just a part of the question or the knowledge fact. 

\subsection{Error Analysis}
Other than errors due to distractions and failed IR, which were around 85\% of the total errors, the errors seen are of four broad categories. 

\textbf{Temporal Reasoning}: In the example \footnote{Predictions are in italics, Correct answers are in Bold.} shown below, even though both the options can be considered as night, the fact that 2:00 AM is more suitable for the bats than 6:00 PM makes it difficult to reason. Such issues accounted for 5\% of the errors. \\
    
    \noindent
     \fbox{\begin{minipage}{19em}
    \textbf{Question:} \textit{ Owls are likely to hunt at? } \\ (A) 3:00 PM (B) \textbf{2:00 AM} (C) 6:00 PM (D)  \textit{7:00 AM}
    
     
    \end{minipage}}\\

    
    
    \textbf{Negation}: In the example shown below, a model is needed which handles negations specifically to reject incorrect options.  Such issues accounted for 1\% of the errors. \\
    
    \noindent
    \fbox{\begin{minipage}{19em}
    \textbf{Question:} \textit{Which of the following is not an input in photosynthesis?} (A) \textit{sunlight} (B) \textbf{oxygen} (C) water (D)  carbon dioxide
    
     
    \end{minipage}}\\
    

    
    
\textbf{Conjunctive Reasoning}: In the example as shown below, each answer options are partially correct as the word ``\textit{ bear}'' is present. Thus a model has to learn whether all parts of the answer are true or not, i.e Conjunctive Reasoning. Logically, all answers are correct, as we can see an ``or'',  but option (A) makes more sense.  Such issues accounted for 1\% of the errors. \\

    \noindent
    \fbox{\begin{minipage}{19em}
    \textbf{Question:} \textit{Some berries may be eaten by } (A) \textbf{ a bear or person} (B) \textit{a bear or shark} (C) a bear or lion (D)  a bear or wolf
    
     
    \end{minipage}}\\
    
\textbf{Qualitative Reasoning}: In the example shown below, each answer options would stop a car but option (D)  is more suitable since it will stop the car quicker. A deeper qualitative reasoning is needed to reject incorrect options.  Such issues accounted for 8\% of the errors. \\

    
    \noindent
     \fbox{\begin{minipage}{19em}
    \textbf{Question:} \textit{ Which of these would stop a car quicker? } (A) a wheel with wet brake pads (B) \textit{a wheel without brake pads} (C) a wheel with worn brake pads (D) \textbf{  a wheel with dry brake pads}
    
     
    \end{minipage}}\\
    
\section{Conclusion}
In this work, we have pushed the current state of the art for the OpenBookQA task using simple techniques and careful selection of knowledge. We have provided two new ways of performing knowledge extraction over a knowledge base for QA and evaluated three ways to perform abductive inference over natural language. All techniques are shown to improve on the performance of the final task of QA, but there is still a long way to reach human performance. 


We analyzed the performance of various components of our QA system. For the natural language abduction, the heuristic technique performs better than the supervised techniques. Our analysis also shows the limitations of BERT based MCQ models, the challenge of learning natural language abductive inference and the multiple types of reasoning required for an OpenBookQA task. Nevertheless, our overall system improves on the state of the art by 11.6\%.


\section{Acknowledgement}
We thank NSF for the grant 1816039 and DARPA for partially supporting this research.

\bibliography{acl2019}

\begin{thebibliography}{34}
\expandafter\ifx\csname natexlab\endcsname\relax\def\natexlab#1{#1}\fi

\bibitem[{Charniak and Goldman(1988)}]{charniak1988logic}
Eugene Charniak and Robert Goldman. 1988.
\newblock A logic for semantic interpretation.
\newblock In \emph{Proceedings of the 26th annual meeting on Association for
  Computational Linguistics}, pages 87--94. Association for Computational
  Linguistics.

\bibitem[{Charniak and Goldman(1989)}]{charniak1989semantics}
Eugene Charniak and Robert~P Goldman. 1989.
\newblock A semantics for probabilistic quantifier-free first-order languages,
  with particular application to story understanding.
\newblock In \emph{IJCAI}, volume~89, pages 1074--1079. Citeseer.

\bibitem[{Choi et~al.(2018)Choi, He, Iyyer, Yatskar, Yih, Choi, Liang, and
  Zettlemoyer}]{choi2018quac}
Eunsol Choi, He~He, Mohit Iyyer, Mark Yatskar, Wen-tau Yih, Yejin Choi, Percy
  Liang, and Luke Zettlemoyer. 2018.
\newblock Quac: Question answering in context.
\newblock \emph{arXiv preprint arXiv:1808.07036}.

\bibitem[{Clark et~al.(2016)Clark, Etzioni, Khot, Sabharwal, Tafjord, Turney,
  and Khashabi}]{clark2016combining}
Peter Clark, Oren Etzioni, Tushar Khot, Ashish Sabharwal, Oyvind Tafjord, Peter
  Turney, and Daniel Khashabi. 2016.
\newblock Combining retrieval, statistics, and inference to answer elementary
  science questions.
\newblock In \emph{Thirtieth AAAI Conference on Artificial Intelligence}.

\bibitem[{Demszky et~al.(2018)Demszky, Guu, and
  Liang}]{demszky2018transforming}
Dorottya Demszky, Kelvin Guu, and Percy Liang. 2018.
\newblock Transforming question answering datasets into natural language
  inference datasets.
\newblock \emph{arXiv preprint arXiv:1809.02922}.

\bibitem[{Devlin et~al.(2018)Devlin, Chang, Lee, and
  Toutanova}]{devlin2018bert}
Jacob Devlin, Ming-Wei Chang, Kenton Lee, and Kristina Toutanova. 2018.
\newblock Bert: Pre-training of deep bidirectional transformers for language
  understanding.
\newblock \emph{arXiv preprint arXiv:1810.04805}.

\bibitem[{Gu et~al.(2016)Gu, Lu, Li, and Li}]{P16-1154}
Jiatao Gu, Zhengdong Lu, Hang Li, and Victor~O.K. Li. 2016.
\newblock \href {https://doi.org/10.18653/v1/P16-1154} {Incorporating copying
  mechanism in sequence-to-sequence learning}.
\newblock In \emph{Proceedings of the 54th Annual Meeting of the Association
  for Computational Linguistics (Volume 1: Long Papers)}, pages 1631--1640.
  Association for Computational Linguistics.

\bibitem[{Hobbs(2004)}]{hobbs2004abduction}
Jerry~R Hobbs. 2004.
\newblock Abduction in natural language understanding.
\newblock \emph{Handbook of pragmatics}, pages 724--741.

\bibitem[{Hobbs et~al.(1993)Hobbs, Stickel, Appelt, and
  Martin}]{hobbs1993interpretation}
Jerry~R Hobbs, Mark~E Stickel, Douglas~E Appelt, and Paul Martin. 1993.
\newblock Interpretation as abduction.
\newblock \emph{Artificial intelligence}, 63(1-2):69--142.

\bibitem[{Honnibal and Montani(2017)}]{spacy2}
Matthew Honnibal and Ines Montani. 2017.
\newblock spacy 2: Natural language understanding with bloom embeddings,
  convolutional neural networks and incremental parsing.
\newblock \emph{To appear}.

\bibitem[{Jenkins(1995)}]{jenkins1995open}
Tony Jenkins. 1995.
\newblock \emph{Open Book Assessment in Computing Degree Programmes}.
\newblock Citeseer.

\bibitem[{Joshi et~al.(2017)Joshi, Choi, Weld, and
  Zettlemoyer}]{joshi2017triviaqa}
Mandar Joshi, Eunsol Choi, Daniel Weld, and Luke Zettlemoyer. 2017.
\newblock \href {https://doi.org/10.18653/v1/P17-1147} {Triviaqa: A large scale
  distantly supervised challenge dataset for reading comprehension}.
\newblock In \emph{Proceedings of the 55th Annual Meeting of the Association
  for Computational Linguistics (Volume 1: Long Papers)}, pages 1601--1611.
  Association for Computational Linguistics.

\bibitem[{Khashabi et~al.(2018)Khashabi, Chaturvedi, Roth, Upadhyay, and
  Roth}]{khashabi2018looking}
Daniel Khashabi, Snigdha Chaturvedi, Michael Roth, Shyam Upadhyay, and Dan
  Roth. 2018.
\newblock Looking beyond the surface: A challenge set for reading comprehension
  over multiple sentences.
\newblock In \emph{Proceedings of the 2018 Conference of the North American
  Chapter of the Association for Computational Linguistics: Human Language
  Technologies, Volume 1 (Long Papers)}, volume~1, pages 252--262.

\bibitem[{Khashabi et~al.(2017)Khashabi, Khot, Sabharwal, and
  Roth}]{khashabi2017learning}
Daniel Khashabi, Tushar Khot, Ashish Sabharwal, and Dan Roth. 2017.
\newblock Learning what is essential in questions.
\newblock In \emph{Proceedings of the 21st Conference on Computational Natural
  Language Learning (CoNLL 2017)}, pages 80--89.

\bibitem[{Lai et~al.(2017)Lai, Xie, Liu, Yang, and Hovy}]{lai2017race}
Guokun Lai, Qizhe Xie, Hanxiao Liu, Yiming Yang, and Eduard Hovy. 2017.
\newblock \href {https://doi.org/10.18653/v1/D17-1082} {Race: Large-scale
  reading comprehension dataset from examinations}.
\newblock In \emph{Proceedings of the 2017 Conference on Empirical Methods in
  Natural Language Processing}, pages 785--794. Association for Computational
  Linguistics.

\bibitem[{Landsberger(1996)}]{jlands1996}
J~Landsberger. 1996.
\newblock \href {Http://www.studygs.net/tsttak7.htm.} {Study guides and
  strategies.}

\bibitem[{Mihaylov et~al.(2018)Mihaylov, Clark, Khot, and
  Sabharwal}]{OpenBookQA2018}
Todor Mihaylov, Peter Clark, Tushar Khot, and Ashish Sabharwal. 2018.
\newblock Can a suit of armor conduct electricity? a new dataset for open book
  question answering.
\newblock In \emph{EMNLP}.

\bibitem[{Mitra et~al.(2019)Mitra, Clark, Tafjord, and
  Baral}]{mitra2019declarative}
Arindam Mitra, Peter Clark, Oyvind Tafjord, and Chitta Baral. 2019.
\newblock Declarative question answering over knowledge bases containing
  natural language text with answer set programming.

\bibitem[{Musa et~al.(2018)Musa, Wang, Fokoue, Mattei, Chang, Kapanipathi,
  Makni, Talamadupula, and Witbrock}]{musa2018answering}
Ryan Musa, Xiaoyan Wang, Achille Fokoue, Nicholas Mattei, Maria Chang, Pavan
  Kapanipathi, Bassem Makni, Kartik Talamadupula, and Michael Witbrock. 2018.
\newblock Answering science exam questions using query rewriting with
  background knowledge.
\newblock \emph{arXiv preprint arXiv:1809.05726}.

\bibitem[{Ni et~al.(2018)Ni, Zhu, Chen, and McAuley}]{ni2018learning}
Jianmo Ni, Chenguang Zhu, Weizhu Chen, and Julian McAuley. 2018.
\newblock Learning to attend on essential terms: An enhanced retriever-reader
  model for scientific question answering.
\newblock \emph{arXiv preprint arXiv:1808.09492}.

\bibitem[{Norvig(1983)}]{norvig1983frame}
Peter Norvig. 1983.
\newblock Frame activated inferences in a story understanding program.
\newblock In \emph{IJCAI}, pages 624--626.

\bibitem[{Norvig(1987)}]{norvig1987inference}
Peter Norvig. 1987.
\newblock Inference in text understanding.
\newblock In \emph{AAAI}, pages 561--565.

\bibitem[{Rajpurkar et~al.(2016)Rajpurkar, Zhang, Lopyrev, and
  Liang}]{rajpurkar2016squad}
Pranav Rajpurkar, Jian Zhang, Konstantin Lopyrev, and Percy Liang. 2016.
\newblock \href {https://doi.org/10.18653/v1/D16-1264} {Squad: 100,000+
  questions for machine comprehension of text}.
\newblock In \emph{Proceedings of the 2016 Conference on Empirical Methods in
  Natural Language Processing}, pages 2383--2392. Association for Computational
  Linguistics.

\bibitem[{Reddy et~al.(2018)Reddy, Chen, and Manning}]{reddy2018coqa}
Siva Reddy, Danqi Chen, and Christopher~D Manning. 2018.
\newblock Coqa: A conversational question answering challenge.
\newblock \emph{arXiv preprint arXiv:1808.07042}.

\bibitem[{Richardson et~al.(2013)Richardson, Burges, and
  Renshaw}]{richardson2013mctest}
Matthew Richardson, Christopher~JC Burges, and Erin Renshaw. 2013.
\newblock Mctest: A challenge dataset for the open-domain machine comprehension
  of text.
\newblock In \emph{Proceedings of the 2013 Conference on Empirical Methods in
  Natural Language Processing}, pages 193--203.

\bibitem[{Seo et~al.(2016)Seo, Kembhavi, Farhadi, and
  Hajishirzi}]{seo2016bidirectional}
Minjoon Seo, Aniruddha Kembhavi, Ali Farhadi, and Hannaneh Hajishirzi. 2016.
\newblock Bidirectional attention flow for machine comprehension.
\newblock \emph{arXiv preprint arXiv:1611.01603}.

\bibitem[{Sun et~al.(2018)Sun, Yu, Yu, and Cardie}]{Sun2018ImprovingMR}
Kai Sun, Dian Yu, Dong Yu, and Claire Cardie. 2018.
\newblock Improving machine reading comprehension with general reading
  strategies.
\newblock \emph{CoRR}, abs/1810.13441.

\bibitem[{Tafjord et~al.(2018)Tafjord, Clark, Gardner, Yih, and
  Sabharwal}]{tafjord2018quarel}
Oyvind Tafjord, Peter Clark, Matt Gardner, Wen-tau Yih, and Ashish Sabharwal.
  2018.
\newblock Quarel: A dataset and models for answering questions about
  qualitative relationships.
\newblock \emph{arXiv preprint arXiv:1811.08048}.

\bibitem[{Vaswani et~al.(2017)Vaswani, Shazeer, Parmar, Uszkoreit, Jones,
  Gomez, Kaiser, and Polosukhin}]{vaswani2017attention}
Ashish Vaswani, Noam Shazeer, Niki Parmar, Jakob Uszkoreit, Llion Jones,
  Aidan~N Gomez, {\L}ukasz Kaiser, and Illia Polosukhin. 2017.
\newblock Attention is all you need.
\newblock In \emph{Advances in Neural Information Processing Systems}, pages
  5998--6008.

\bibitem[{Wang et~al.(2018)Wang, Singh, Michael, Hill, Levy, and
  Bowman}]{wang2018glue}
Alex Wang, Amapreet Singh, Julian Michael, Felix Hill, Omer Levy, and Samuel~R
  Bowman. 2018.
\newblock Glue: A multi-task benchmark and analysis platform for natural
  language understanding.
\newblock \emph{arXiv preprint arXiv:1804.07461}.

\bibitem[{Wilensky(1983)}]{wilensky1983planning}
Robert Wilensky. 1983.
\newblock Planning and understanding: A computational approach to human
  reasoning.

\bibitem[{Wilensky et~al.(2000)Wilensky, Chin, Luria, Martin, Mayfield, and
  Wu}]{wilensky2000berkeley}
Robert Wilensky, David~N Chin, Marc Luria, James Martin, James Mayfield, and
  Dekai Wu. 2000.
\newblock The berkeley unix consultant project.
\newblock In \emph{Intelligent Help Systems for UNIX}, pages 49--94. Springer.

\bibitem[{Zellers et~al.(2018)Zellers, Bisk, Schwartz, and
  Choi}]{zellers2018swag}
Rowan Zellers, Yonatan Bisk, Roy Schwartz, and Yejin Choi. 2018.
\newblock Swag: A large-scale adversarial dataset for grounded commonsense
  inference.
\newblock \emph{arXiv preprint arXiv:1808.05326}.

\bibitem[{Zhang et~al.(2018)Zhang, Dai, Toraman, and Song}]{zhang2018kg}
Yuyu Zhang, Hanjun Dai, Kamil Toraman, and Le~Song. 2018.
\newblock Kg\^{} 2: Learning to reason science exam questions with contextual
  knowledge graph embeddings.
\newblock \emph{arXiv preprint arXiv:1805.12393}.

\end{thebibliography}
\bibliographystyle{acl_natbib}

\appendix

\end{document}